\documentclass[sigconf]{acmart}

\usepackage{multirow}
\usepackage{booktabs}
\usepackage{utfsym}
\usepackage{diagbox}
\usepackage{ctable}
\usepackage{pifont}
\usepackage{ulem}
\usepackage{graphicx}
\usepackage{multirow,multicol}
\usepackage{subfigure}
\usepackage{color}
\usepackage{bm}
\usepackage{epsfig}
\usepackage{rotating}
\usepackage{epstopdf}

\usepackage{algorithmic}
\usepackage{algorithm}

\allowdisplaybreaks[4]

\renewcommand\footnotetextcopyrightpermission[1]{} 
\begin{document}

\title{Graph Unlearning with Efficient Partial Retraining}

\author{Jiahao Zhang, Lin Wang, Shijie Wang, Wenqi Fan}
\affiliation{
   \institution{The Hong Kong Polytechnic University}
   \country{Kowloon, Hong Kong SAR}
}
\email{{tony-jiahao.zhang, comp-lin.wang, shijie.wang}@connect.polyu.hk, wenqifan03@gmail.com}

\renewcommand{\shortauthors}{Jiahao Zhang, Wenqi Fan, Lin Wang, Shijie Wang.}

\acmConference[WWW '24] {Companion Proceedings of the ACM Web Conference 2024}{May 13--17, 2024}{Singapore, Singapore.}

\begin{abstract}

Graph Neural Networks (GNNs) have achieved remarkable success in various real-world applications. 
However, GNNs may be trained on undesirable graph data, which can degrade their performance and reliability.
To enable trained GNNs to efficiently unlearn unwanted data, a desirable solution is retraining-based graph unlearning, which partitions the training graph into subgraphs and trains sub-models on them, allowing fast unlearning through partial retraining. 
However, the graph partition process causes information loss in the training graph, resulting in the low model utility of sub-GNN models. 
In this paper, we propose GraphRevoker, a novel graph unlearning framework that better maintains the model utility of unlearnable GNNs. 
Specifically, we preserve the graph property with graph property-aware
sharding and effectively aggregate the sub-GNN models for prediction with graph contrastive
sub-model aggregation. 
We conduct extensive experiments to demonstrate the superiority of our proposed approach. 

\end{abstract}



\maketitle

\section{Introduction}

\begin{figure}[t]
    \centering
     \vskip -0.45in
    \includegraphics[width=0.9\linewidth]{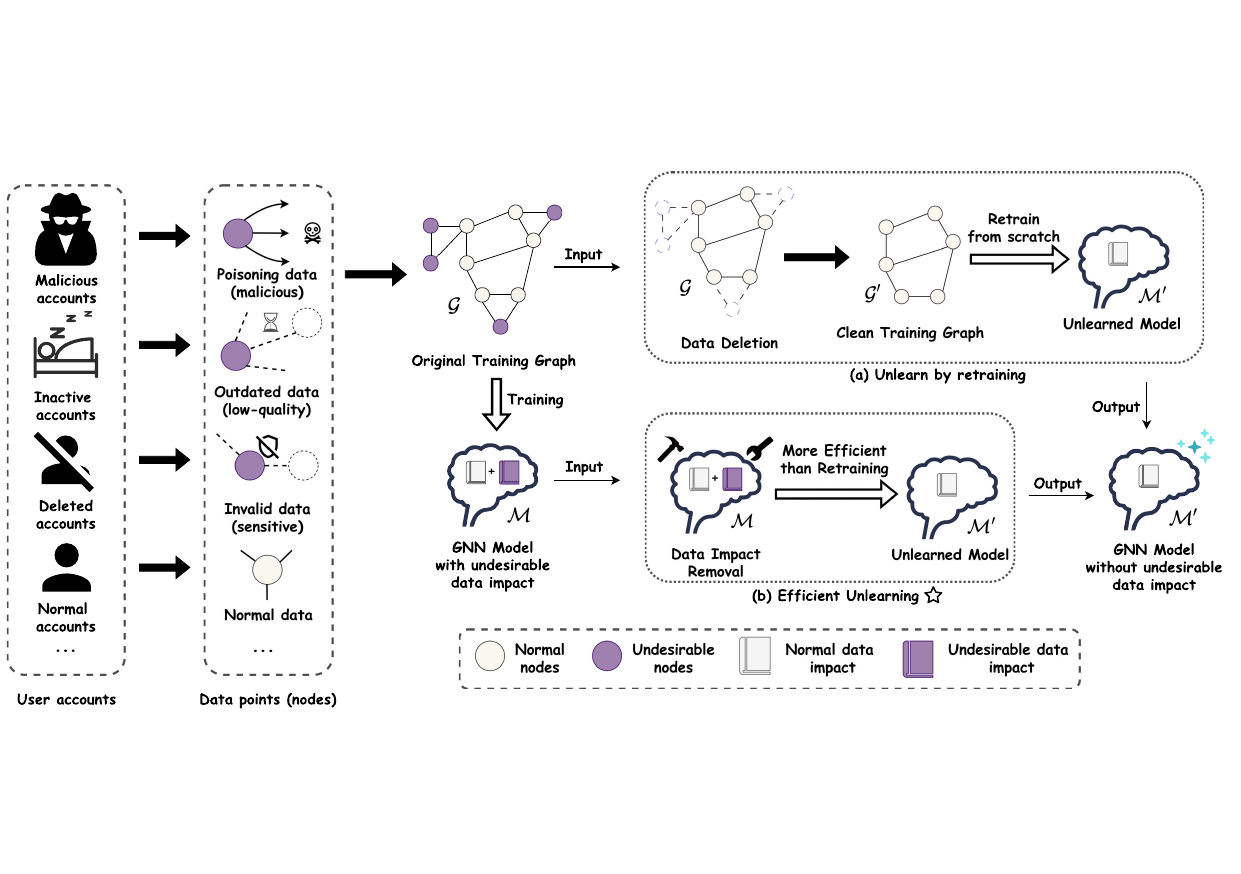}
    \vskip -0.55in
    \caption{An example of graph unlearning in a social network.}
    \vskip -0.3in
    \label{fig:motivation}
\end{figure}

As one of the crucial data representations, graphs are used to describe data with complex relations between objects. Recently, Graph Neural Networks (\textbf{GNNs}) have achieved remarkable success in learning from graph data in various real-world applications. 

Despite the aforementioned success of GNNs, the concerns about undesirable data posing detrimental effects on GNNs are rising. 
For instance, \emph{malicious data}~\cite{chen2022knowledge, fan2023jointly}, \emph{low-quality data}~\cite{fan2022graph, wang2023fast}, and \emph{sensitive data}~\cite{chen2021graph, chen2022recommendation} are threatening GNNs' safety, prediction performance, and compliance to "the right to be forgotten" (as shown in Fig.~\ref{fig:motivation}), which necessities the development of \textbf{graph unlearning}. 

To efficiently remove the effect of undesirable data from trained GNN models, recent years have witnessed two separate lines of research: 1) \emph{approximate unlearning} and 2) \emph{retraining-based unlearning}. 
Approximate unlearning relies on directly manipulating the model parameters to erase the effects of unwanted data points~\cite{chenggnndelete, wu2023gif}, which sacrifices the reliability of the removal, as discussed in previous literatures~\cite{bourtoule2021machine, thudi2022necessity}. 
On the other hand, retraining approaches partition the training graph into small subgraphs and train disjoint sub-models on them, allowing the model owner only to retrain a small sub-model to remove the effect of some bad data. 
Despite their desirable security properties~\cite{bourtoule2021machine, thudi2022necessity}, these unlearning approaches may destroy the graph structure and label semantic during the partitioning process, degrading the model utility of sub-model GNNs.

In this work, our research objective revolves around 1) preserving the desirable property of retraining-based unlearning and 2) significantly improving the model utility of sub-GNN models in this paradigm. 
To achieve this goal, we systematically review the limitations of prior works in graph unlearning (as discussed in Section~\ref{sec:relatedwork}), and then introduce a novel graph unlearning framework, namely \textbf{GraphRevoker}, equipped with \emph{graph property-aware sharding} and \emph{graph contrastive sub-model aggregation}. Our contributions can be summarized as follows: 
\textbf{1)} We propose the \emph{graph property-aware sharding} module to preserve the sub-GNN models' prediction performance by keeping the structural and semantic properties in the training graph. 
\textbf{2)} To effectively leverage the disjoint sub-models for prediction, we propose the \emph{graph contrastive sub-model aggregation} module, a lightweight GNN ensemble network, empowered by local-local structural reconstruction and local-global contrastive learning. 
\textbf{3)} Extensive experiments 
are conducted to illustrate the model utility and unlearning efficiency of the proposed method.

\section{Problem Formulation}

\noindent \textbf{Notations.} 
In general, a graph can be denoted as $\mathcal{G}=(\mathcal{V}, \mathcal{E})$, where $\mathcal{V}=\{u_1, \cdots, u_N\}$ denotes the node set, and the edge set $\mathcal{E}$ is represented by adjacency matrix $\boldsymbol{A}$.  
We define the diagonal degree matrix $\boldsymbol{D}$ as $\boldsymbol{D}_{i, i}=deg(u_i)=\sum_{j=1}^N{\boldsymbol{A}_{i, j}}$. 
For node classification, the labels of nodes can be represented as $\mathcal{Y}=[y_1, \cdots, y_{N}]$, in which $y_i\in\{\ell_1, \cdots, \ell_C\}$ denotes the $C$ different categories.

\noindent \textbf{Problem Definition.}
\label{sec:def}
With the concerns on malicious and low-quality graph data, given a trained GNN model $\mathcal{F}_{\boldsymbol{\theta}}$, the goal of graph unlearning is to eliminate the impact of an undesirable subset of training data $\mathcal{D}^{-}$ from $\mathcal{F}_{\boldsymbol{\theta}}$ while preserving its model utility. 
Specifically, in the context of node classification, the undesirable subset with size $t$ can be represented as a set of undesirable nodes $\mathcal{D}^{-}=\{v_{j_1}, \cdots, v_{j_t}\} \subset \mathcal{V}$.

Since undesirable knowledge from $\mathcal{D}^{-}$ has been encoded into the parameters of $\mathcal{F}_{\boldsymbol{\theta}}$, \emph{retraining-based graph unlearning} aims to obtain an unlearned GNN model $\mathcal{F}_{\boldsymbol{\theta}_u'}$, which is equivalent to the GNN model $\mathcal{F}_{\boldsymbol{\theta}_r'}$ that \emph{has never been trained} on $\mathcal{D}^{-}$. 
This definition of graph unlearning can be formalized as follows:
\begin{align*}
       \text{\textbf{Retraining:} }~~~~ 
       &\mathcal{G} \xrightarrow[]{\text{data removal}} \mathcal{G} / \mathcal{D}^{-}\xrightarrow[]{\text{retrain}} \mathcal{F}_{\boldsymbol{\theta}_r'}; \\
       \text{\textbf{Unlearning:} }~~~~  &\mathcal{G} \xrightarrow[]{\text{train}} \mathcal{F}_{\boldsymbol{\theta}} \xrightarrow[]{\text{unlearn} (\mathcal{D^{-}})} \mathcal{F}_{\boldsymbol{\theta}_u'},
\end{align*}
where $\mathcal{F}_{\boldsymbol{\theta}_u'}$ and $\mathcal{F}_{\boldsymbol{\theta}_r'}$ are expected to be equivalent. 

In graph unlearning, we also aim to achieve two goals: \emph{1) Model Utility}: After unlearning data points $\mathcal{D}^{-}$, the latest model $\mathcal{F}_{\boldsymbol{\theta}_u'}$ should have comparable performance in comparison with retrained model $\mathcal{F}_{\boldsymbol{\theta}_r'}$; \emph{2) Unlearning Efficiency}: Compared to retraining from scratch, the unlearning process is expected to be more efficient. 

\section{Related Work}
\label{sec:relatedwork}

\begin{table}
\centering
\caption{Characteristics of different unlearning frameworks.}
\vskip -0.15in
\scalebox{0.75}{
\begin{tabular}{l|c|ccc}
\specialrule{.1em}{.05em}{.05em} 
\multirow{2}{*}{\diagbox{\textbf{Aspect}}{\textbf{Method}}}                  & \multirow{2}{*}{Approximate\cite{wu2023gif, chenggnndelete}}    & \multicolumn{3}{c}{Retraining-based}      \\
                        &  & SISA\cite{bourtoule2021machine}  & GraphEraser\cite{chen2021graph} & Ours \\
\hline
Accurate Removal               & \ding{55}              & \ding{51}     & \ding{51}           & \ding{51}    \\
Structural Preservation & \ding{51}              & \ding{55}     & \ding{51}           & \ding{51}    \\
Semantic Preservation  & \ding{51}              & \ding{51}     & \ding{55}           & \ding{51}    \\
Effective Ensemble     &  N/A          & \ding{55}     & \ding{55}           & \ding{51} \\   
\specialrule{.1em}{.05em}{.05em} 
\end{tabular}
}
\vskip -0.25in
\label{tab:model_adv}
\end{table}

In this section, we briefly review the state-of-the-art approaches in graph unlearning, and further discuss their limitations compared with our proposed framework. Additional related works can be found in Section~\ref{sec:append_related}. 

\noindent \textbf{Approximate Unlearning}. To avoid the costly retraining process when unlearning undesirable data, approximate approaches directly manipulate the parameters of GNNs, and alleviate the impact of the unwanted data. 
For instance, GIF~\cite{wu2023gif} develops a novel graph influence function tailored to GNNs and updates the parameters with the gradients from the influence function, while GNNDelete~\cite{chenggnndelete} inserts a trainable delete operator between each GNN layer and optimizing the parameters of the delete operator to achieve unlearning. 
However, recent literature~\cite{bourtoule2021machine, thudi2022necessity} finds these methods do not guarantee the accurate removal of the deleted data, which necessitates the existence of retraining in unlearning frameworks.

\noindent \textbf{Efficient Retraining}. 
Regarding both accurate removal and unlearning efficiency, SISA~\cite{bourtoule2021machine} has introduced the retraining-based paradigm, randomly partitioning the training graph into small subgraphs, thereby training sub-models on them. Thus, exact data impact removal can be achieved by only retraining a sub-model. 
Later on, GraphEraser~\cite{chen2021graph} adapts the SISA~\cite{bourtoule2021machine} paradigm to the context of graphs by introducing a clustering-based partitioning strategy and learnable sub-model aggregation, enhancing the model prediction performance while still allowing efficient and exact unlearning. 

Despite the desirable removal guarantee, retraining-based graph unlearning still faces challenges in sub-optimal model utility, due to the graph structural loss in the partition stage. 
It is true that clustering-based partitions~\cite{chen2021graph} provide a remedy for this issue, but cluster-based partitions tend to put nodes with the same label into one subgraph, sacrificing the semantic balance between different submodels. 
In addition, the lack of expressive sub-GNN aggregators also deteriorates the model utility problem in retraining approaches. 

To this end, we propose GraphRevoker, which achieves efficient and accurate data removal, while preserving both the structural and semantic information in the graph partition phase. We also introduce a powerful contrastive learning empowered aggregation module to ensemble the subgraph GNNs. The advantages of our proposed model can be seen in Table.~\ref{tab:model_adv}.

\section{Proposed Method}\label{sec:methodlogy}

\begin{figure*}[t]
    \centering
    \vskip -0.45in
    \includegraphics[width=0.9\linewidth]{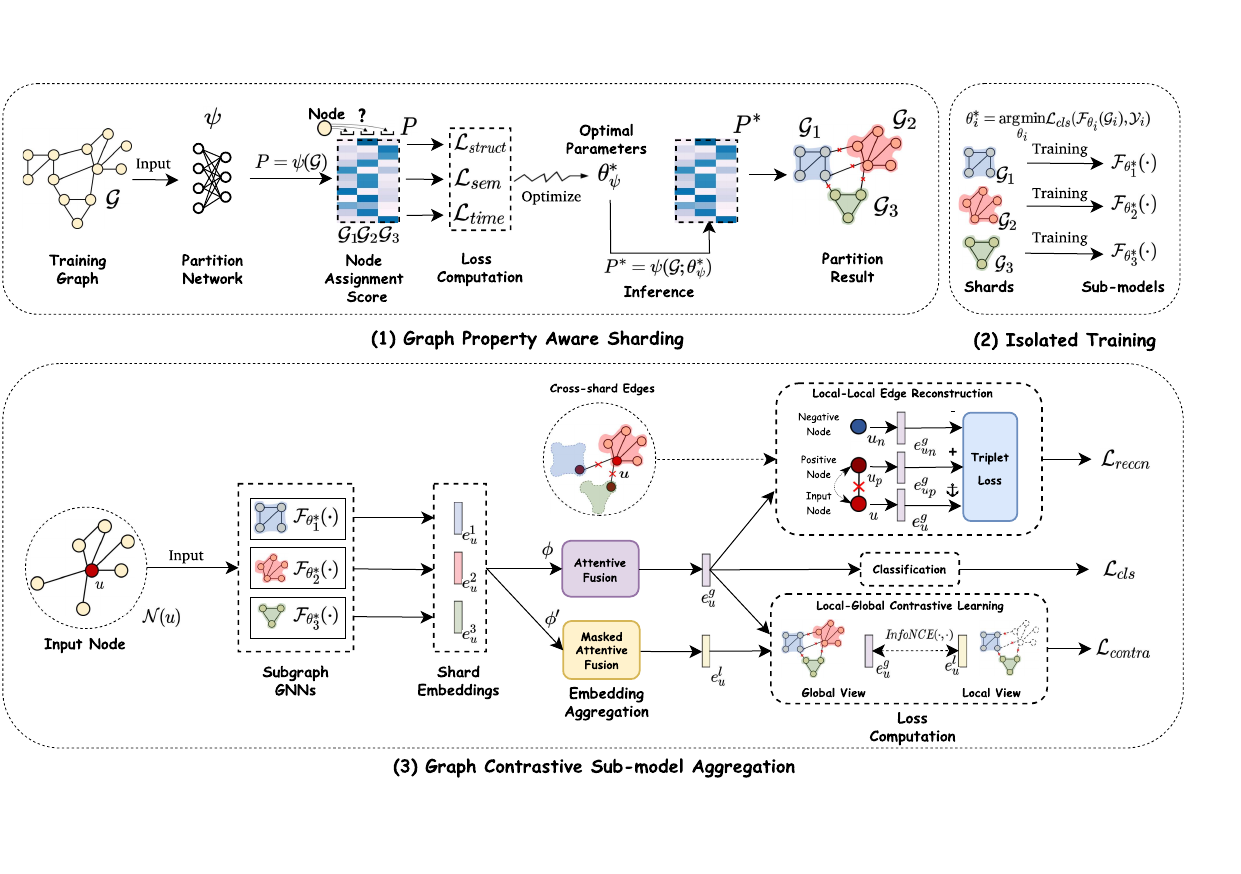}
    \vskip -0.62in
    \caption{Illustration of the proposed framework.} 
    \vskip -0.15in
    \label{fig:frame}
\end{figure*}

To endow GNNs with efficient and exact unlearning capability, GraphRevoker follows the widely adopted SISA framework, which mainly contains three stages: 1) subgraph partition, 2) isolated training, and 3) sub-model aggregation, as shown in Fig.~\ref{fig:frame}. In both partitioning and aggregation stages, we introduce innovative designs to preserve the model utility of the framework. 
After the aforementioned steps, we can apply GraphRevoker to make predictions by aggregating all the predictions from subgraph GNN models, or efficiently unlearn data points by partially retraining the affected sub-models.

\subsection{Graph Property-Aware Sharding}
\label{sec:method_part}
To preserve model utility and unlearning efficiency in the partition phase, we first formulate the unlearning goals in Section~\ref{sec:def} into three reachable optimization objectives, and then solve them with an effective neural framework to give desirable graph partitions. 
In comparison to previous random~\cite{bourtoule2021machine} and clustering-based~\cite{chen2021graph} partition methods, our framework preserves both graph structure and label semantics, resulting in stronger sub-GNN models. 

\noindent \textbf{Unlearning Time.} The efficiency of retraining-based unlearning mainly depends on the time cost of retraining corresponding sub-models, which relies on two key factors: 1) the probability of retraining a specific sub-model and 2) the time cost of retraining that sub-model. Clearly, the first factor relates to the number of nodes in a subgraph, and the second factor lies behind the cost of message-passing, which is proportional to the number of edges~\cite{zhang2024linear}. Let $S$ denote the number of partitioned subgraphs, and $\mathcal{P} = \{\mathcal{V}_1,\cdots,\mathcal{V}_S\}$ denotes node partition results, satisfying $\forall\mathcal{V}_i\in\mathcal{V}$ and $\forall i\neq j, \mathcal{V}_i\cap\mathcal{V}_j=\emptyset$. We propose the unlearning time objective as the expectation of retraining time as follows:
\begin{equation}
\label{eq:ltime}
    \mathcal{L}_{time} = \sum_{i=1}^SPr(\mathcal{V}_i)\cdot\text{Cost}(\mathcal{V}_i) \approx\sum_{i=1}^S\frac{|\mathcal{V}_i|}{|\mathcal{V}|}|\mathcal{E}_i|.
\end{equation}

\noindent \textbf{Graph Structure Preservation.} 
Edges that connect nodes in different subgraphs are inevitably removed during the partition phase. However, to maintain the model utility of sub-GNNs, preserving the original graph's structure is crucial. Prior works in graph unlearning~\cite{chen2021graph, chen2022recommendation} achieve structural preservation by using balanced K-Means~\cite{chen2021graph} or balanced Label Propagation~\cite{chen2022recommendation} in an unsupervised setting, which cannot directly protect the graph structure. To address this issue, we propose a principled and supervised objective to quantize the structural loss in the graph partitions, namely the normalized edge-cut ($Ncut$) objective, counting the number of dropped edges between different subgraphs as follows:
\begin{equation}
\label{eq:lstruct}
    \mathcal{L}_{struct} = Ncut(\mathcal{G}_1,\mathcal{G}_2,\cdots,\mathcal{G}_S) = \sum_{k=1}^{S} \frac{cut(\mathcal{G}_k,\bar{\mathcal{G}}_k)}{\sum_{u_i\in \mathcal{G}_k} deg(u_i) }, 
\end{equation}
where $cut(\mathcal{G}_k,\bar{\mathcal{G}}_k)$ denotes the edges between subgraph $\mathcal{G}_k$ and the remaining part of the training graph.

\noindent \textbf{Label Semantic Preservation.} Besides the structural destruction in the partition phase, the label distribution of the training graph is also perturbed, which may result in biased and less generalizable subgraph models. This phenomenon is more severe in clustering-based partitions~\cite{chen2021graph, chen2022recommendation}, which tend to assign nodes with similar labels into one subgraph. To overcome this problem, we propose an objective to capture the richness of label semantics in each subgraph with the entropy of its label distribution. 

Let $c_{\ell_j}(\mathcal{V}_i) = |\{ u_k|u_k \in \mathcal{V}_i, \mathcal{Y}_{k} = \ell_j \}|$ denote the number of nodes annotated with label $\ell_j$ in the $i$-th shard. We have the discrete label distribution $\boldsymbol{d}(\mathcal{V}_i)$ for the $i$-th subgraph, where $Pr(\ell_i) = \frac{c_{\ell_i}(\mathcal{V}_i)}{|\mathcal{V}_i|}$. Therefore, we propose the entropy-based semantic preservation objective as follows:
\begin{equation}\label{eq:lsem}
\mathcal{L}_{sem} = \frac{1}{S}\sum_{i=1}^{S} \text{Entropy}\left[\boldsymbol{d}(\mathcal{V}_i)\right] = \frac{1}{S}\sum_{i=1}^{S} {\sum_{j=1}^C{-\log[\frac{c_{\ell_j}(\mathcal{V}_i)}{|\mathcal{V}_i|}]\frac{c_{\ell_j}(\mathcal{V}_i)}{|\mathcal{V}_i|}}}.
\end{equation}

\noindent \textbf{Differentiable Graph Partition Framework.} To partition the training graph $\mathcal{G}$ into disjoint subgraphs while minimizing objectives Eqs.~\eqref{eq:ltime}~-~\eqref{eq:lsem}, the main challenge lies behind optimization. As graph partition is a combinatorial optimization problem and cannot be solved in polynomial time, we relax the graph partition into a continuous form and solve it with a neural network. 

To make graph partitioning continuous, we represent the partition result with a soft node assignment matrix $\boldsymbol{P}\in \mathbb{R}^{N\times S}$, where $\boldsymbol{P}_{i, j}\in [0, 1]$ and $\sum\boldsymbol{P}_{i, :} = 1$. The node assignment matrix can be computed with a graph partition network $\psi$ parameterized with $\boldsymbol{\theta}_\psi$, which contains GNN layers and a softmax output layer, transforming the node representations into the assignment matrix. Therefore, we can learn how to give effective graph partitions by training network $\psi$ with the following loss function:
\begin{align}\label{eq:partition_loss}
    \mathcal{L}_{part} &= \mathcal{L}_{time} + \mathcal{L}_{struct} + \mathcal{L}_{sem} + \frac{1}{2}\gamma\lVert\boldsymbol{\theta}_{\psi}\rVert_2^2, \\
    \label{eq:timeloss_wrt_p}
    \text{s.t.}\quad \mathcal{L}_{time} &= \frac{1}{|\mathcal{V}|}\sum_{i=1}^S (\boldsymbol{1}^T\boldsymbol{P}_{:, i})\sum_{reduce}[(\boldsymbol{P}_{:, i}\boldsymbol{P}_{:, i}^T)\odot\boldsymbol{A}],\\
    \mathcal{L}_{struct} &= \sum_{reduce} {[\boldsymbol{P}\cdot diag^{-1}(\boldsymbol{1}^T\boldsymbol{D}\boldsymbol{P})](1-\boldsymbol{P}^T)\odot \boldsymbol{A}},\\
    \label{eq:semloss_wrt_p}
    \mathcal{L}_{sem} &= \frac{1}{S}\sum_{i=1}^S\sum_{j=1}^C-log\left(\frac{\boldsymbol{P}^T_{:, i}\boldsymbol{Y}_{:, j}}{\boldsymbol{P}_{:,i}^T\boldsymbol{1}}\right)\frac{\boldsymbol{P}^T_{:, i}\boldsymbol{Y}_{:, j}}{\boldsymbol{P}_{:,i}^T\boldsymbol{1}},
\end{align}
where $\boldsymbol{1}$ denotes 1-valued column vector, and $\boldsymbol{Y}\in\{0,1\}^{|\mathcal{V}|\times C}$ denotes the one-hot label matrix. After training $\psi$ on the training graph with $\mathcal{L}_{part}$, we can use this network to infer desirable partitions. Please refer to Section~\ref{sec:append_loss} in our supplementary material for further explanations of the partition loss function.

\subsection{Graph Contrastive Sub-model Aggregation}
\label{aggr}

After acquiring the subgraph partition $\mathcal{P} = \{\mathcal{V}_1, \cdots, \mathcal{V}_S\}$, we can train $S$ isolated submodels $\mathcal{F}_{\boldsymbol{\theta}_1},\cdots, \mathcal{F}_{\boldsymbol{\theta}_S}$ on these disjoint subgraphs. Thus, unlearning can be efficiently achieved by only retraining a single sub-model affected by unwanted data. 
However, unlearning efficiency is not the only goal of retraining-based unlearning, and another challenge lies behind the difficulty of leveraging the weak sub-models to obtain an accurate prediction. Though previous works have explored some straightforward solutions, including mean averaging~\cite{bourtoule2021machine}, weighted averaging~\cite{chen2021graph}, and attention mechanism~\cite{chen2022recommendation}, it is still an open question on how to utilize the graph structures in the sub-model aggregation phase. 

In this work, we develop a contrastive learning framework to learn an effective aggregator to ensemble the weak sub-GNN models. The aggregator only contains a few parameters and can be trained efficiently with only a small subset of training nodes, namely $\mathcal{U} = \{u_{j_1},\cdots, u_{j_{M}}\}\subset\mathcal{V}, |\mathcal{U}|\ll|\mathcal{V}|$. 

\noindent \textbf{Attentive Fusion.} When making predictions with the trained sub-models, every sub-model $\mathcal{F}_{\theta_i}(\cdot)$ generates a node embedding matrix $\boldsymbol{E}^i\in\mathbb{R}^{|\mathcal{U}|\times d}$, where row vector $\boldsymbol{e}^i_u$ represents the embedding of node $u\in\mathcal{U}$. Towards a better global prediction, we first align embeddings from different feature spaces with a learnable linear projection, and then fuse all the embeddings with an attention mechanism. Let $\bar{\boldsymbol{E}}\in\mathbb{R}^{|\mathcal{V}|\times d}$ denote the fused embedding matrix, and $\bar{\boldsymbol{e}}_u$ denote a row vector in $\bar{\boldsymbol{E}}$, we define the attentive fusion as follows:
\begin{align}\label{eq:attn_score_avg}
    \alpha^i_u &= \frac{\exp(\boldsymbol{w}^T\text{ReLU}(\boldsymbol{W}_i\boldsymbol{e}_u^i+\boldsymbol{b}_i))}{\sum_{j=1}^S\exp(\boldsymbol{w}^T\text{ReLU}(\boldsymbol{W}_j\boldsymbol{e}_u^j+\boldsymbol{b}_j))};\quad \bar{\boldsymbol{e}}_u = \frac{1}{S}\sum_{i=1}^S\alpha^i_u\boldsymbol{e}^i_u,
\end{align}
where $\{\boldsymbol{W}_1, \cdots,\boldsymbol{W}_S\}$, $\{\boldsymbol{b}_1,\cdots,\boldsymbol{b}_S\}$ and $\boldsymbol{w}$ are trainable parameters, and $\alpha^i_u$ is the attention score for sub-model $i$ and node $u$. 

\noindent \textbf{Local-global Contrastive Loss.} In unlearning frameworks based on partial retraining, the output from sub-models contains local knowledge of subgraphs in $\mathcal{G}$, while the aggregation result is expected to incorporate a full knowledge of $\mathcal{G}$. 
Inspired by previous Graph Contrastive Learning (GCL) approaches~\cite{zhu2021graph, wu2022disentangled}, we find it natural to leverage the local views of a node to enhance its global aggregation result $\bar{\boldsymbol{e}}^u$. 

Specifically, we regard the fully aggregated embedding $\bar{\boldsymbol{e}}_u$ as a global view of node $u$, while we generate a local view $\tilde{\boldsymbol{e}}_u$ by randomly deactivating some sub-model attention scores in Eq.~\eqref{eq:attn_score_avg} (i.e., $\tilde{\boldsymbol{e}}_u=\frac{S}{\lVert\boldsymbol{m}\rVert_1}\sum_{i=1}^S\boldsymbol{m}_i\alpha^i_u\boldsymbol{e}^i_u$, where $\boldsymbol{m}$ is a random 0-1 mask). Thus, two different views of the same node $(\bar{\boldsymbol{e}}_u, \tilde{\boldsymbol{e}}_u)$ are expected to be close, and the representation from another random node $v\neq u$ should be pulled apart. The local-global contrastive loss is formulated with the InfoNCE objective as follows:
\begin{align}
    \mathcal{L}_{contra} &= \frac{1}{|\mathcal{U}|}\sum_{u \in \mathcal{U}} \mathcal{L}_{InfoNCE}(\bar{\boldsymbol{e}}_u, \tilde{\boldsymbol{e}}_u), \\
    \label{infonce}\ \ \mathcal{L}_{InfoNCE}(\bar{\boldsymbol{e}}_{u}, \tilde{\boldsymbol{e}}_{u}) &= -\log\frac{ e^{\phi(\bar{\boldsymbol{e}}_u, \tilde{\boldsymbol{e}}_u)/\tau} + e^{\phi(\bar{\boldsymbol{e}}_u, \bar{\boldsymbol{e}}_{v})/\tau} + e^{\phi(\bar{\boldsymbol{e}}_u, \tilde{\boldsymbol{e}}_{v})/\tau}}{e^{\phi(\bar{\boldsymbol{e}}_u, \tilde{\boldsymbol{e}}_u)/\tau}},
\end{align}
where $\phi(\cdot,\cdot)$ denotes the cosine similarity function, and $\tau$ denotes the softmax temperature. 
This formulation not only considers both inter-view negative pairs $(\bar{\boldsymbol{e}}_u, \bar{\boldsymbol{e}}_{v})$ and intra-view negative pairs $(\bar{\boldsymbol{e}}_u, \tilde{\boldsymbol{e}}_{v})$, but also requires zero external data augmentation, resulting in expressive representations and efficient computations. 

\noindent \textbf{Local-local Reconstruction Loss.} In graph partitioning and isolated training, the links between subgraphs are dropped to ensure each node's impact only exists in one sub-model, which allows unlearning by only retraining one specific sub-model. 
Nevertheless, the dropped edges could include useful structural information of training graph $\mathcal{G}$. To address this problem, we propose the local-local reconstruction loss to restore the knowledge of the previously ignored edges as follows:
\begin{equation}
    \mathcal{L}_{recon} = \frac{1}{|\mathcal{U}|}\sum_{u\in\mathcal{U}}{\max\{\phi(\bar{\boldsymbol{e}}_u, \bar{\boldsymbol{e}}_{v^+}) - \phi(\bar{\boldsymbol{e}}_u, \bar{\boldsymbol{e}}_{v^-}) + 1 ,0\}},
\end{equation}
where the positive sample $\bar{\boldsymbol{e}}_{v^+}$ is sampled from the neighbors of $u$ in other subgraphs, and the negative sample $\bar{\boldsymbol{e}}_{v^{-}}$ does not have any connection with $u$. 

\noindent \textbf{Optimization.} Our aggregation module learns to fuse sub-GNN models by minimizing the following training objective:
\begin{align}
    \mathcal{L}_{aggr} = {\mathcal{L}_{cls} + \mathcal{L}_{contra} + \mathcal{L}_{recon} + \frac{1}{2}\gamma'||\boldsymbol{\Theta}||_2^2},
\end{align}
where $\boldsymbol{\Theta}$ denotes the trainable parameters of the linear projection and attention mechanism. After unlearning each data point, the aggregation module must be retrained to ensure an accurate removal. Fortunately, we find this module needs 10 to 20 epochs to be trained, only introducing a small computational overhead. 

\section{Experiment}

\noindent \textbf{Datasets.} We evaluate GraphRevoker on four real-world graph datasets, including \emph{Cora}~\cite{yang2016revisiting}, \emph{Citeseer}~\cite{yang2016revisiting}, \emph{LastFM-Asia}~\cite{rozemberczki2020characteristic}, and \emph{Flickr}~\cite{zeng2019graphsaint}. 
For the first three datasets, we randomly divide nodes into train/val/test sets with a 0.7/0.2/0.1 ratio. For Flickr, we use the pre-defined dataset splits in PyG. More implementation details and additional experiments are presented in Section~\ref{sec:append_setting} and Section~\ref{sec:append_exp}.

\noindent \textbf{Model Performance.} We have evaluated all the unlearning frameworks on the inductive node classification task. The model utility and unlearning efficiency results can be found in Tables.~\ref{tab:acc} ~-~\ref{tab:time}, and we draw the following conclusions: 1) GraphRevoker witnessed the state-of-the-art model performance among all the three efficient unlearning frameworks in both model utility and unlearning efficiency; 2) Despite the strong model utility of Retrain, its efficiency is far worse than efficient unlearning approaches, which is unacceptable in large-scale settings; 3) The propose GraphRevoker even outperforms SISA in efficiency evaluations, which can be attributed to our retraining-time-aware design in the partition module and our parallelized sub-model retraining implementations.

\begin{table}[t]
\setlength\tabcolsep{3pt}
\caption{The comparison results of model utility in F1-score.}
\vskip -0.15in
\label{tab:acc}
\scalebox{0.6}{
\begin{tabular}{c|l|ccccc}
\hline
\textbf{Datasets}              & \multicolumn{1}{c|}{\textbf{Methods}} & \textbf{GAT} & \textbf{GCN} & \textbf{SAGE} & \textbf{APPNP} & \textbf{JKNet} \\ \hline
\multirow{4}{*}{Cora}          & Retrain                              & \underline{83.95±0.67}   & \underline{84.46±0.51}   & \underline{82.84±0.73}    & \underline{83.66±0.71}     & \underline{84.52±0.54}     \\
                               & SISA                                 & 67.47±4.43   & 55.02±1.06   & 58.82±4.87    & 66.64±4.61     & 68.21±4.30     \\
                               & GraphEraser                          & 75.70±0.33   & 71.27±2.99   & 72.10±0.58    & 68.84±2.56     & 69.94±3.12     \\
                               & \textbf{GraphRevoker}                & \textbf{76.94±1.17}   & \textbf{78.75±0.47}   & \textbf{75.94±1.39}    & \textbf{72.95±0.61}     &\textbf{70.90±1.79}     \\ \hline
\multirow{4}{*}{Citeseer}      & Retrain                              & \underline{73.37±0.83}   & \underline{72.94±0.63}   & \underline{73.17±0.75}    & \underline{71.04±0.83}     & \underline{73.24±1.01}     \\
                               & SISA                                 & 55.90±5.28   & 69.01±3.33   & 63.32±4.72    & 62.99±5.31     & 65.32±4.58     \\
                               & GraphEraser                          & 60.53±5.10   & 69.59±1.39   & 67.16±1.85    & 68.36±3.21     & 68.71±2.20     \\
                               & \textbf{GraphRevoker}                & \textbf{70.27±0.80}   & \textbf{72.79±1.04}   & \textbf{69.35±1.42}    & \textbf{70.09±1.67}     & \textbf{70.62±1.45}     \\ \hline
\multirow{4}{*}{LastFM  Asia}  & Retrain                              & \underline{85.29±0.53}   & \underline{84.40±0.73}   & \underline{82.18±0.34}    & \underline{83.66±0.71}     & \underline{85.36±0.46}     \\
                               & SISA                                 & 75.77±0.44   & 75.46±0.52   & 74.42±0.61    & 76.06±0.35     & 75.67±0.38     \\
                               & GraphEraser                          & 68.12±2.10   & 64.77±0.96   & 64.52±1.91    & 71.55±0.90     & 69.87±0.77     \\
                               & \textbf{GraphRevoker}                & \textbf{76.43±0.93}   & \textbf{76.60±0.63}   & \textbf{75.14±0.62}    & \textbf{77.56±0.71}     & \textbf{76.20±0.77}     \\ \hline
\multirow{4}{*}{Flickr}        & Retrain                              & \underline{49.36±0.64}   & \underline{49.93±0.51}   & \underline{49.41±1.39}    & \underline{48.50±0.80}     & \underline{49.61±1.67}     \\
                               & SISA                                 & 42.81±0.86   & 46.01±1.24   & 46.92±1.24    & 46.72±1.02     & 44.38±0.95     \\
                               & GraphEraser                          & 43.82±1.68   & 46.42±1.10   & 47.52±0.62    & 45.96±1.66     & 44.32±1.55     \\
                               & \textbf{GraphRevoker}                & \textbf{45.19±1.42}   & \textbf{48.35±0.63}   & \textbf{48.09±0.11}    & \textbf{47.36±0.82}     & \textbf{46.16±1.33}     \\ \hline
\end{tabular}
}
\end{table}

\begin{table}[t]
\setlength\tabcolsep{3pt}
\vskip -0.15in
\caption{The comparison results of the time-cost of unlearning the undesirable data points ($\mathcal{D}^-$) (unit: second[s]).}
\vskip -0.15in
\label{tab:time}
\scalebox{0.6}{
\begin{tabular}{c|c|l|rrrrr}
\hline
\textbf{\begin{tabular}[c]{@{}c@{}}Unlearned \\ Nodes\end{tabular}} & \textbf{Datastets}             & \multicolumn{1}{c|}{\textbf{Methods}} & \multicolumn{1}{c}{\textbf{GAT}} & \multicolumn{1}{c}{\textbf{GCN}} & \multicolumn{1}{c}{\textbf{SAGE}} & \multicolumn{1}{c}{\textbf{APPNP}} & \multicolumn{1}{c}{\textbf{JKNet}} \\ \hline
\multirow{16}{*}{0.5\%}                                            & \multirow{4}{*}{Cora}          & Retrain                              & 30.71              & 24.93                 & 17.55                 & 26.26                  & 24.42                  \\
                                                                    &                                & SISA                                 & 7.79                             & 5.04                             & 5.43                              & 5.74                               & 6.41                               \\
                                                                    &                                & GraphEraser                          & 10.37                            & 7.77                             & 6.40                              & 8.52                               & 8.55                               \\
                                                                    &                                & \textbf{GraphRevoker}                & \textbf{4.72}                    & \textbf{4.65}                    & \textbf{3.10}                     & \textbf{4.23}                      & \textbf{4.88}                \\ \cline{2-8} 
                                                                    & \multirow{4}{*}{Citeseer}      & Retrain                              & 36.56               & 30.73                 & 18.80               & 29.20                   & 32.10               \\
                                                                    &                                & SISA                                 & 9.52                             & 7.30                             & 4.80                              & 7.57                               & 6.16                               \\
                                                                    &                                & GraphEraser                          & 16.39                            & 10.51                            & 9.89                              & 12.90                              & 12.70                              \\
                                                                    &                                & \textbf{GraphRevoker}                & \textbf{6.25}                    & \textbf{6.05}                    & \textbf{3.85}                     & \textbf{5.32}                     & \textbf{5.87}                   \\ \cline{2-8} 
                                                                    & \multirow{4}{*}{LastFM   Asia} & Retrain                              & 106.86              & 88.57                 & 73.96                & 96.63                 & 100.31               \\
                                                                    &                                & SISA                                 & 31.59                            & 46.33                            & 30.58                             & 55.04                              & 51.81                              \\
                                                                    &                                & GraphEraser                          & 35.51                            & 46.95                            & 36.50                             & 50.18                              & 52.84                              \\
                                                                    &                                & \textbf{GraphRevoker}                & \textbf{12.64}                  & \textbf{29.49}                   & \textbf{16.46}                     & \textbf{28.57}                  & \textbf{27.99}                 \\ \cline{2-8} 
                                                                    & \multirow{4}{*}{Flickr}        & Retrain                              & 177.11               & 138.67              & 108.06                & 155.04               & 151.35                 \\
                                                                    &                                & SISA                                 & 139.68                           & 110.88                           & 72.29                             & 123.16                             & 115.53                             \\
                                                                    &                                & GraphEraser                          & 152.04                           & 120.76                           & 87.86                             & 130.22                             & 127.81                             \\
                                                                    &                                & \textbf{GraphRevoker}                & \textbf{57.37}                  & \textbf{53.69}                   & \textbf{22.75}                     & \textbf{41.74}                   & \textbf{55.11}                  \\ \hline

\end{tabular}
}
\vskip -0.1in
\end{table}

\section{Conclusion}
\label{sec:conclusion}

Graph unlearning, which involves removing the impact of undesirable data points from trained GNNs, is significant in various real-world scenarios. In this work, we propose a novel graph unlearning framework, which fully unleashes the potential of retraining-based unlearning in graphs and GNNs to achieve accurate, model utility preserving, and efficient unlearning. 
To alleviate the model degradation attributed to the partition and isolated training process, we introduce two main contributions: a neural graph partition network and a graph contrastive sub-model ensemble module.
In our future works, we plan to explore the effectiveness of GraphRevoker on more settings and downstream tasks in graph mining.

\balance

\bibliographystyle{ACM-Reference-Format}
\bibliography{references_short}

\clearpage
\nobalance
\appendix
\section{Supplementary Material}
\subsection{Evaluation Settings}\label{sec:append_setting}
\begin{table}[htbp]
\vskip -0.1in
\caption{Dataset statistics.}
\vskip -0.1in
\label{tab:dataset}
\scalebox{0.8}{
\begin{tabular}{c|cccc}
\hline
\textbf{Dataset}     & \textbf{\# Nodes} & \textbf{\# Edges} & \textbf{\# Classes} & \textbf{Type} \\ \hline
\textbf{Cora}~\cite{yang2016revisiting}        & 2,708             & 5,429             & 7                   & Citation      \\
\textbf{Citeseer}~\cite{yang2016revisiting}    & 3,327             & 4,732             & 6                   & Citation      \\
\textbf{LastFM-Asia}~\cite{rozemberczki2020characteristic} & 7,264             & 55,612            & 18                  & Social Network        \\
\textbf{Flickr}~\cite{zeng2019graphsaint}      & 89,250            & 899,756           & 7                   & Web Image     \\
\hline
\end{tabular}
}
\vskip -0.1in
\end{table}

\subsubsection{\textbf{Dataset Statistics.}} We evaluate GraphRevoker on four real-world graph datasets from various origins, including \emph{Cora}, \emph{Citeseer}, \emph{LastFM-Asia}, and \emph{Flickr}, which are widely used to evaluate the performance of GNN models. The statistic of the datasets is shown in Table.~\ref{tab:dataset}.

\subsubsection{\textbf{General Settings.}} We select \emph{GAT}~\cite{velivckovic2018graph}, \emph{GCN}~\cite{kipf2016semi}, \emph{SAGE}~\cite{hamilton2017inductive}, \emph{APPNP}~\cite{gasteiger2018predict}, and \emph{JKNet}~\cite{xu2018representation} as our base models to evaluate the unlearning frameworks. All the GNN models except \emph{JKNet} include two message-passing layers and an MLP classifier, and all the \emph{JKNet}s contain three message-passing layers. The embedding size for GNN models is set to 64. The partition module is optimized with an AdamW optimizer with lr=1e-3 and weight\_decay=1e-5. Specifically, the partition network $\psi$ is trained in 10 to 30 epochs until convergence, and weights for $\mathcal{L}_{time}$ and $\mathcal{L}_{sem}$ are both set to 1e-3. The aggregation module is optimized with an AdamW optimizer with lr=0.01 and weight\_decay=1e-5. The weights for both auxiliary loss functions in aggregator training ($\mathcal{L}_{contra}$ and $\mathcal{L}_{recon}$) are set to 1e-4. For compared baselines (\emph{SISA} and \emph{GraphEraser}), we followed their official implementations and settings, and carefully tuned their hyperparameters based on their suggested hyperparameter search space (e.g., the learning rate of the aggregator).

\label{sec:util_setting}
\subsubsection{\textbf{Model Utility Settings.}} For all unlearning methods except \emph{Retrain}, the number of shards is fixed to 20. Each experiment is conducted 10 times, and we report both the average and standard deviation of the results. All the GNN models are trained for 100 epochs except on the \emph{Flickr} dataset, where we train the GNN models for 20 epochs in \emph{SISA}, \emph{GraphEraser} and our \emph{GraphRevoker} to avoid over-fitting. For the number of samples to train the learnable aggregator ($|\mathcal{U}|$), we follow the settings of \emph{GraphEraser}, which selects 1000 samples in datasets with small or medium sizes (\emph{Cora}, \emph{Citerseer} and \emph{LastFM Asia}), and selected 10\% of the training nodes in larger datasets (\emph{Flickr}).

\subsubsection{\textbf{Unlearning Efficiency Settings.}} We randomly select 0.5\% nodes from the node-set $\mathcal{V}$ as the undesirable data points $\mathcal{D}^-$, and report the total time cost, including retraining sub-models affected by undesirable data points and reconstructing the aggregator. The settings of GNN models and the aggregator are the same as we mentioned in Section~\ref{sec:util_setting}.
With a single RTX 3090 GPU, 112$\times$ Intel Xeon Gold 6238R CPU cores, and 10 sub-processes, we run the experiments and present the unlearning efficiency in Table.~\ref{tab:time}.

\subsection{Additional Experiments}\label{sec:append_exp}

\begin{figure}[htbp]
\vskip -0.1in
\centering
\subfigure[GAT - Cora]{
\begin{minipage}[t]{0.30\linewidth}
\centering
\includegraphics[width=\linewidth]{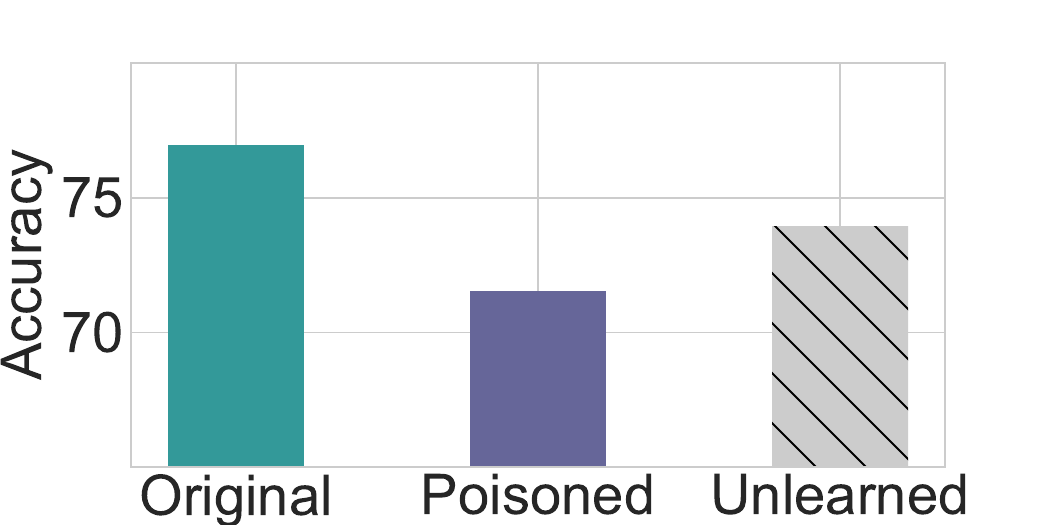}
\end{minipage}
}
\subfigure[GCN - Cora]{
\begin{minipage}[t]{0.30\linewidth}
\centering
\includegraphics[width=\linewidth]{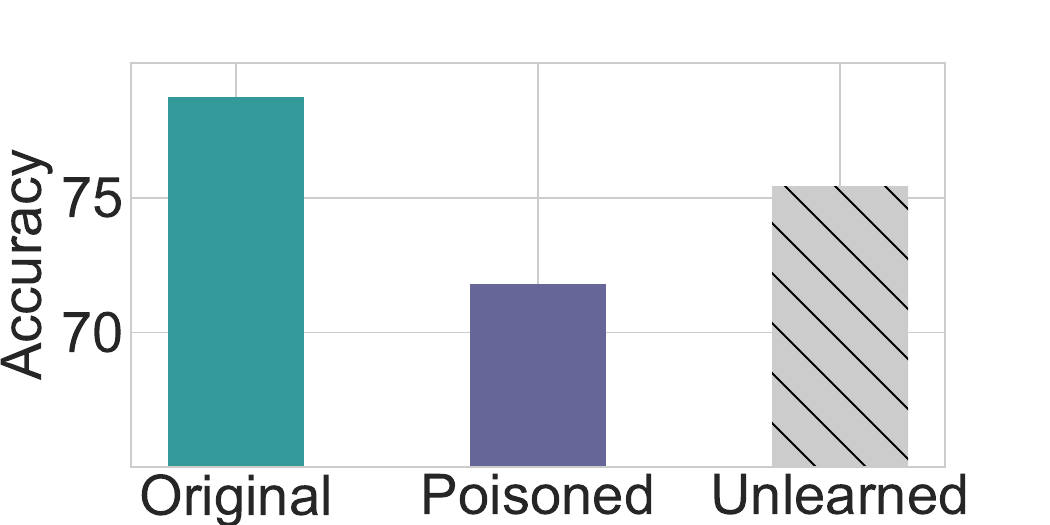}
\end{minipage}
}
\subfigure[APPNP - Cora]{
\begin{minipage}[t]{0.30\linewidth}
\centering
\includegraphics[width=\linewidth]{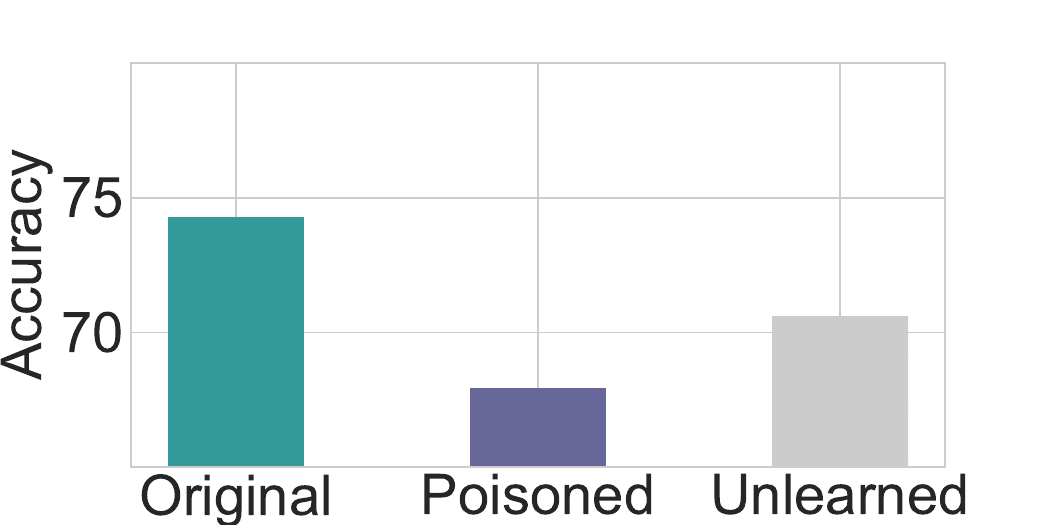}
\end{minipage}
}

\subfigure[GAT - Citeseer]{
\begin{minipage}[t]{0.30\linewidth}
\centering
\includegraphics[width=\linewidth]{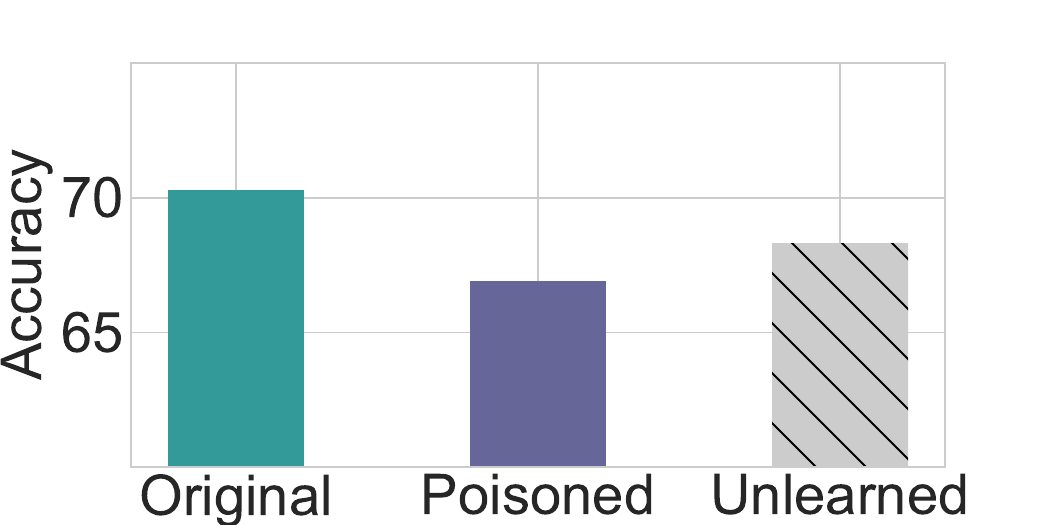}
\end{minipage}
}
\subfigure[GCN - Citeseer]{
\begin{minipage}[t]{0.30\linewidth}
\centering
\includegraphics[width=\linewidth]{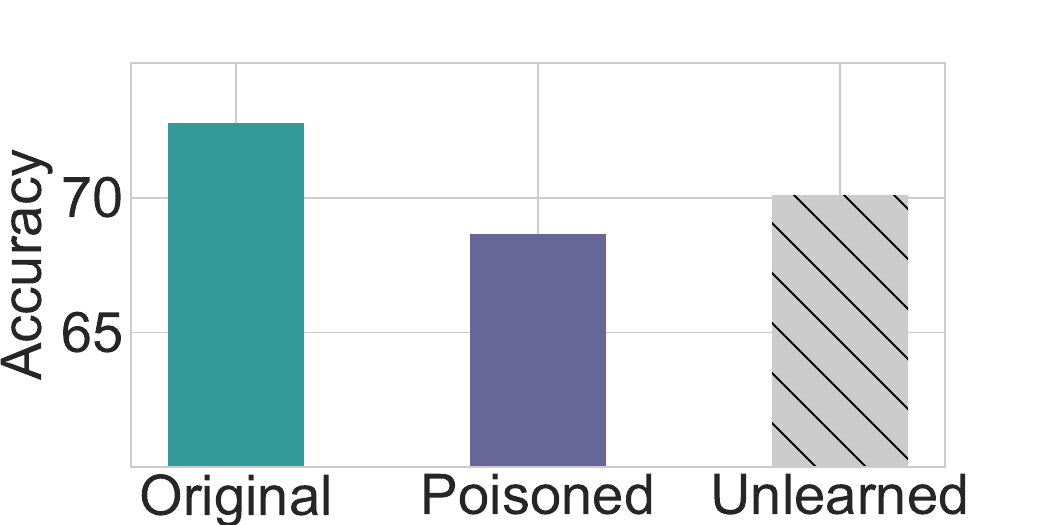}
\end{minipage}
}
\subfigure[APPNP - Citeseer]{
\begin{minipage}[t]{0.30\linewidth}
\centering
\includegraphics[width=\linewidth]{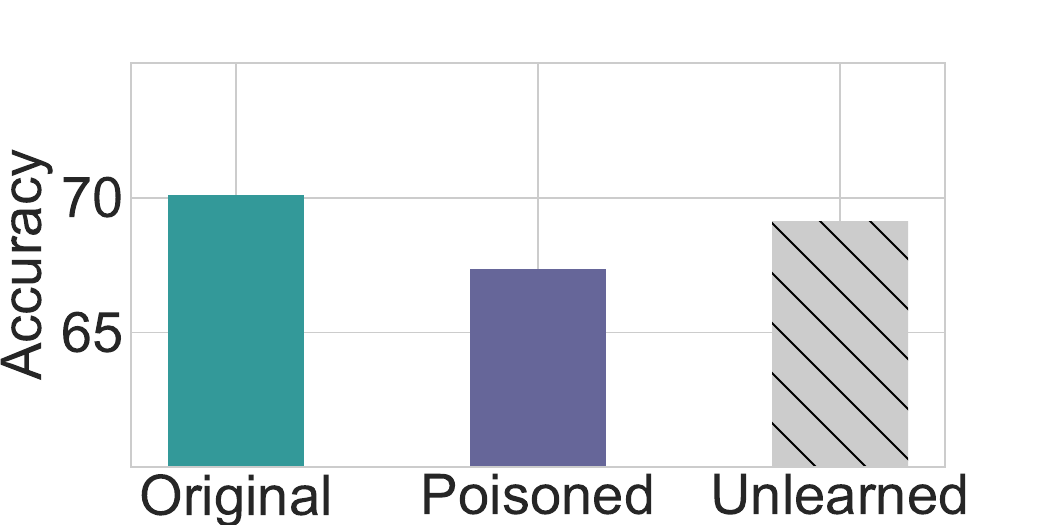}
\end{minipage}
}

\centering
\vskip -0.1in
\caption{ The unlearning power of our proposed framework. This figure shows the F1-score (\%) on Cora and Citeseer of the original model, the model affected by low-quality data, and the model unlearned the low-quality data. }
\label{fig:attack}
\vskip -0.2in
\end{figure}

\subsubsection{\textbf{Unlearning Power.}} 
We verify the unlearning power of GraphRevoker by enabling models affected by low-quality data points to regain their utility through unlearning undesirable data points. 
In a real-world setting, low-quality data (e.g., outdated user-item interactions in recommendations) can be regarded as data points with noisy labels, so here we adopt the random attack~\cite{ma2021graph, fan2023untargeted} to inject 100 nodes with wrong labels on Cora and Citeseer. 
We follow their settings and add 10 random edges for each injected node. In Figure.~\ref{fig:attack}, we show the F1-score of the original GNN models and models affected by noisy data in (a), (b), (d), and (e), and then we unlearn the injected negative data points with GraphRevoker and record the model utility in (c) and (f). 
As illustrated in Figure.~\ref{fig:attack}, we can find that the GNN models poisoned by low-quality data suffer a dramatic performance decrease, and after unlearning the injected noisy data points, the model utility could be partly regained. 
Though the model utility cannot be fully recovered as the noisy data also degrades the graph sharding module, our proposed graph unlearning framework could efficiently unlearn undesirable data, and the recovered model utility is comparable.

\begin{table}[ht]
\setlength\tabcolsep{2pt}
\vskip -0.1in
\caption{Ablation study for the partition module.}
\vskip -0.15in
\label{tab:ablation2}
\scalebox{0.6}{
\begin{tabular}{c|l|cccccc}
\hline
\multirow{2}{*}{\textbf{Datasets}} & \multicolumn{1}{c|}{\multirow{2}{*}{\textbf{Design}}} & \multicolumn{2}{c}{\textbf{GAT}}     & \multicolumn{2}{c}{\textbf{GCN}}    & \multicolumn{2}{c}{\textbf{APPNP}}  \\ \cline{3-8} 
                                   & \multicolumn{1}{c|}{}                                 & F1-Score          & Time     & F1-Score          & Time    & F1-Score          & Time    \\ \hline
\multirow{4}{*}{\textbf{Cora}}     & \textbf{GraphRevoker}                                 & \textbf{76.94±1.17} & \textbf{4.72}  & \textbf{77.66±1.60} & \textbf{4.65} & \textbf{75.35±1.31} & \textbf{4.23} \\
                                   & \textbf{w/o $\mathcal{L}_{sem}$}                      & 71.99±1.60          & 4.77           & 73.82±0.68          & 4.6           & 69.54±1.26          & 4.16          \\
                                   & \textbf{w/o $\mathcal{L}_{time}$}                       & 75.89±1.21          & 4.92           & 75.70±1.73          & 4.73          & 71.85±1.63          & 4.3           \\
                                   & \textbf{w/o $\mathcal{L}_{time} +   \mathcal{L}_{sem}$} & 70.94±1.50          & 5.07           & 71.66±0.82          & 4.93          & 72.64±1.75          & 4.47          \\ \hline
\multirow{4}{*}{\textbf{Flickr}}   & \textbf{GraphRevoker}                                 & \textbf{45.19±1.42} & \textbf{57.37} & \textbf{48.35±0.63} & 53.69         & \textbf{47.36±0.82} & 41.74         \\
                                   & \textbf{w/o $\mathcal{L}_{sem}$}                      & 44.46±0.33          & 57.94          & 47.64±1.11          & 52.23         & 46.68±1.01          & 40.48         \\
                                   & \textbf{w/o $\mathcal{L}_{time}$}                       & 44.54±0.89          & 58.44          & 48.00±1.20          & 54.91         & 47.19±0.88          & 42.54         \\
                                   & \textbf{w/o $\mathcal{L}_{time} +   \mathcal{L}_{sem}$} & 44.09±0.36          & 59.98          & 47.88±0.91          & 55.22         & 46.09±1.39          & 44.41         \\ \hline
\end{tabular}
}
\vskip -0.1in
\end{table}

\begin{table}[ht]
\vskip 0.05in
\caption{Ablation study for the aggregation module.}
\vskip -0.15in
\label{tab:ablation3}
\scalebox{0.6}{
\begin{tabular}{c|l|ccc}
\hline
\textbf{Datasets}                & \multicolumn{1}{c|}{\textbf{Design}} & \textbf{GAT} & \textbf{GCN} & \textbf{APPNP} \\ \hline
\multirow{4}{*}{\textbf{Cora}}   & \textbf{GraphRevoker}               & \textbf{76.94±1.17}   & \textbf{78.75±0.47}   & \textbf{74.27±1.16}\\
                                 & \textbf{w/o} $\mathcal{L}_{contra}$                & 76.03±1.02   & 78.56±0.90   & 73.75±1.41     \\
                                 & \textbf{w/o} $\mathcal{L}_{recon}$                     & 75.70±1.50   & 78.34±1.21   & 74.13±0.31     \\
                                 & \textbf{w/o} $\mathcal{L}_{contra} + \mathcal{L}_{recon}$                           & 74.91±2.19   & 78.15±1.06   & 73.39±1.19     \\ \hline
\multirow{4}{*}{\textbf{Flickr}} &\textbf{GraphRevoker}               & \textbf{45.19±1.42}   & \textbf{48.35±0.63}   & \textbf{47.36±0.82} \\
                                 & \textbf{w/o} $\mathcal{L}_{contra}$                & 44.21±0.72   & 47.97±0.49   & 47.12±0.83     \\
                                 & \textbf{w/o} $\mathcal{L}_{recon}$                     & 44.48±1.55   & 48.20±0.82   & 47.00±1.27     \\
                                 & \textbf{w/o} $\mathcal{L}_{contra} + \mathcal{L}_{recon}$                           & 43.86±1.20   & 47.47±1.07   & 46.30±1.15     \\ \hline
\end{tabular}
}
\vskip -0.1in
\end{table}
\subsubsection{\textbf{Ablation Study.}} As shown in Tables~\ref{tab:ablation2}-\ref{tab:ablation3}, we present the performance of GraphRevoker with different designs of training objectives in both partition and aggregation modules. We find that ablating any part of our design will result in a model performance drop. This illustrates the effectiveness of different components in our proposed GraphRevoker framework.

\subsection{Explanations of the Partition Loss Function}\label{sec:append_loss}
In Section~\ref{sec:method_part}, we directly present the loss function $\mathcal{L}_{part}$ of the differentiable graph partition framework in Eq.~\eqref{eq:partition_loss} to Eq.~\eqref{eq:semloss_wrt_p} w.r.t. the soft node assignment matrix $\boldsymbol{P}$. However, it is still unclear why Eq.~\eqref{eq:timeloss_wrt_p} to Eq.~\eqref{eq:semloss_wrt_p} based on $\boldsymbol{P}$ are sufficient to represent the concepts (e.g., edge cuts, node counts, etc.) in Eq.~\eqref{eq:ltime} to Eq.~\eqref{eq:lsem}. In this section, we supplement the derivation details of the partition training objective, illustrating the intuition behind the computations in Eq.~\eqref{eq:timeloss_wrt_p} to Eq.~\eqref{eq:semloss_wrt_p}. 

The general idea of understanding the soft node assignment matrix $\boldsymbol{P}$ is to interpret it from a probabilistic perspective. Specifically, $\boldsymbol{P}_{i, j}$ denotes the probability of assigning node $u_i$ to subgraph $\mathcal{V}_j$, so we can estimate the expectation of the number of nodes, number of edges, and edge cuts in each subgraph. When inferring actual graph partitions with trained partition network $\psi$, we directly assign nodes to the subgraph with the largest probability (i.e., assigning node $u_i$ to subgraph indexed $\mathop{\boldsymbol{\arg\max}}_{j}\left(\boldsymbol{P}_{i, j}\right)$ ). 

\subsubsection{\textbf{The Unlearning Time Objective.}} To compute Eq.~\eqref{eq:ltime} with the soft assignment matrix $\boldsymbol{P}\in\mathbb{R}^{N\times S}$, the key problem is representing the number of nodes $|\mathcal{V}_i|$ and edges $|\mathcal{E}_i|$ in each partitioned subgraph $\mathcal{V}_i$. It is straightforward to count the nodes in each subgraph, which is 
\begin{equation}\label{eq:node_cnt_wrt_p}
\mathbb{E}|\mathcal{V}_i|=\sum_{j=1}^{N}\boldsymbol{P}_{j, i}\times 1=\boldsymbol{1}^T\boldsymbol{P}_{:, i},   
\end{equation}
where $\boldsymbol{1}$ denotes the 1-valued column vector. To count the edges in the partitioned subgraph, we have to traverse all the edges $(u_j, u_k)$ in $\mathcal{E}$ and compute the expectation as follows:
\begin{equation}
    \mathbb{E}|\mathcal{E}_i| = \sum_{(u_j, u_k)\in\mathcal{E}}Pr(u_j, u_k \in \mathcal{V}_i)\times 1=\sum_{\boldsymbol{A}_{j, k}=1} \boldsymbol{P}_{j, i}\boldsymbol{P}_{k, i}.
\end{equation}
However, computing $|\mathcal{E}_i|$ with loop structures is computationally inefficient, so we further give a vectorized form, which is
\begin{equation}\label{eq:edge_cnt_wrt_p}
    \mathbb{E}|\mathcal{E}_i| = \sum_{reduce}[(\boldsymbol{P}_{:, i}\boldsymbol{P}_{:, i}^T)\odot\boldsymbol{A}],
\end{equation}
where $\odot$ denotes point-wise multiplication, and $\sum_{reduce}$ denotes the reduce-sum operator. Thus, with Eq.~\eqref{eq:node_cnt_wrt_p} and Eq.~\eqref{eq:edge_cnt_wrt_p}, we can recover the proposed unlearning time loss in Eq.~\eqref{eq:timeloss_wrt_p} as follows:
\begin{align*}
    \mathcal{L}_{time} &= \sum_{i=1}^S\frac{|\mathcal{V}_i|}{|\mathcal{V}|}|\mathcal{E}_i| \\
    &\approx \frac{1}{|\mathcal{V}|}\sum_{i=1}^S\mathbb{E}|\mathcal{V}_i|\mathbb{E}|\mathcal{E}_i| \\
    &= \frac{1}{|\mathcal{V}|}\sum_{i=1}^S (\boldsymbol{1}^T\boldsymbol{P}_{:, i})\sum_{reduce}[(\boldsymbol{P}_{:, i}\boldsymbol{P}_{:, i}^T)\odot\boldsymbol{A}].
\end{align*}

\subsubsection{\textbf{The Structure Preservation Objective.}} 
The normalized edge cut objective in Eq.~\eqref{eq:lstruct} includes the edge cut in the numerator and the subgraph degree summation in the denominator. First, we show the computation of the edge cut based on the node assignment probabilities:
\begin{equation}\label{eq:ecut}
\begin{aligned}
        \mathbb{E}[cut(\mathcal{G}_i, \bar{\mathcal{G}}_i)] &= \sum_{(u_j, u_k)\in\mathcal{E}} Pr(u_j \in \mathcal{V}_i) Pr(u_k \notin \mathcal{V}_i)\times 1 \\
        &= \sum_{\boldsymbol{A}_{j, k}=1} \boldsymbol{P}_{j, i}(1 - \boldsymbol{P}_{k, i})\\
        &= \sum_{reduce}[\boldsymbol{P}_{:, i}(1 - \boldsymbol{P}_{:, i})^T]\odot \boldsymbol{A}. 
\end{aligned}
\end{equation}

The expected node degree in a subgraph can be computed as follows:
\begin{equation}\label{eq:edeg}
\begin{aligned}
    \mathbb{E}\left[\sum_{u_j\in\mathcal{V}_i}deg(u_j)\right] = \sum_{j=1}^N \boldsymbol{P}_{j, i}deg(u_j) = \boldsymbol{1}^T\boldsymbol{D}\boldsymbol{P}_{:, i}.        
\end{aligned}
\end{equation}
Therefore, we can combine previous Eq.~\eqref{eq:ecut} and Eq.~\eqref{eq:edeg} to recover the $Ncut$ objective in the structure preservation loss as follows:
\begin{equation*}
\begin{aligned}
    \mathcal{L}_{sem} &= \sum_{k=1}^{S} \frac{cut(\mathcal{G}_k,\bar{\mathcal{G}}_k)}{\sum_{u_i\in \mathcal{G}_k} deg(u_i) } \\
    &\approx\sum_{k=1}^{S} \frac{\mathbb{E}\left[cut(\mathcal{G}_k,\bar{\mathcal{G}}_k)\right]}{\mathbb{E}\left[\sum_{u_i\in \mathcal{G}_k} deg(u_i)\right]} \\
    &= \sum_{k=1}^S \frac{\sum_{reduce}[\boldsymbol{P}_{:, k}(1 - \boldsymbol{P}_{:, k})^T]\odot\boldsymbol{A}}{\boldsymbol{1}^T\boldsymbol{D}\boldsymbol{P}_{:, k}} \\
    &= \sum_{reduce}\left[\sum_{k=1}^S\frac{\boldsymbol{P}_{:, k}(1 - \boldsymbol{P}_{:, k})^T}{\boldsymbol{1}^T\boldsymbol{D}\boldsymbol{P}_{:, k}}\right] \odot\boldsymbol{A} \\
    &= \sum_{reduce} {[\boldsymbol{P}\cdot diag^{-1}(\boldsymbol{1}^T\boldsymbol{D}\boldsymbol{P})](1-\boldsymbol{P}^T)\odot \boldsymbol{A}}.
\end{aligned}    
\end{equation*}
\subsubsection{\textbf{The Semantic Preservation Objective.}} To recover the semantic preservation loss in Eq.~\eqref{eq:semloss_wrt_p}, the computation of the number of nodes $|\mathcal{V}_i|$ is shown in Eq.~\eqref{eq:node_cnt_wrt_p}. The only problem is counting the number of nodes annotated with label $\ell_j$ (i.e., $c_{\ell_j}(\mathcal{V}_i)$), which is computed as follows:
\begin{equation}\label{eq:label_cnt}
    \mathbb{E}[c_{\ell_j}(\mathcal{V}_i)] = \sum_{\mathcal{Y}_k=\ell_j}\boldsymbol{P}_{k, i}\times 1 = \sum_{i=1}^N \boldsymbol{P}_{k, i} \times\boldsymbol{Y}_{k, j} = \boldsymbol{P}^T_{:, k}\boldsymbol{Y}_{:, j}.
\end{equation}
Therefore, with Eq.~\eqref{eq:node_cnt_wrt_p} and Eq.~\eqref{eq:label_cnt}, we have the following derivations for the semantic preservation loss in Eq.~\eqref{eq:semloss_wrt_p}:
\begin{align*}
    \mathcal{L}_{sem} &= \frac{1}{S}\sum_{i=1}^{S} {\sum_{j=1}^C{-\log[\frac{c_{\ell_j}(\mathcal{V}_i)}{|\mathcal{V}_i|}]\frac{c_{\ell_j}(\mathcal{V}_i)}{|\mathcal{V}_i|}}} \\
    &\approx \frac{1}{S}\sum_{i=1}^{S} {\sum_{j=1}^C{-\log[\frac{\mathbb{E}[c_{\ell_j}(\mathcal{V}_i)]}{\mathbb{E}|\mathcal{V}_i|}]\frac{\mathbb{E}[c_{\ell_j}(\mathcal{V}_i)]}{\mathbb{E}|\mathcal{V}_i|}}} \\
    &= \frac{1}{S}\sum_{i=1}^S\sum_{j=1}^C-log\left(\frac{\boldsymbol{P}^T_{:, i}\boldsymbol{Y}_{:, j}}{\boldsymbol{P}_{:,i}^T\boldsymbol{1}}\right)\frac{\boldsymbol{P}^T_{:, i}\boldsymbol{Y}_{:, j}}{\boldsymbol{P}_{:,i}^T\boldsymbol{1}}.
\end{align*}

\subsection{Extended Related Works}\label{sec:append_related}
\subsubsection{\textbf{Machine Unlearning}.} 
Machine Unlearning is a novel concept that denotes removing the impact of some undesirable data points from trained ML models. 
Pioneering efforts in this domain, including statistical query unlearning~\cite{cao2015towards} and certified removal~\cite{guo2020certified}, have provided effective and provable data impact removal solutions for simple ML models (e.g., Logistic Regression). 
Nevertheless, these methods cannot be simply adapted to more complex models, such as Deep Neural Networks (DNNs). 
To mitigate this, researchers have introduced approximate unlearning strategies aimed at removing the effect of adverse data in DNNs, which estimates the influence of these data on parameters and manipulates the corresponding parameters~\cite{golatkar2020eternal, warnecke2023machine}. 
In retrospect, approximate unlearning methods only ensure that the impact of data is alleviated, falling short of achieving exact data impact removal. 
To address this problem, Bourtoule et al. proposed the SISA~\cite{bourtoule2021machine} framework to achieve model-agnostic and exact unlearning via data partitioning and partial retraining. 
Owing to its flexibility and reliable removal guarantee, SISA and its variants have made great progress in machine unlearning tailored to image recognition~\cite{yan2022arcane} and decision trees~\cite{brophy2021machine}. 

Despite the aforementioned success of machine unlearning approaches, graph unlearning for graph neural networks (GNNs) is still a less explored problem. It is also non-trivial to adapt the existing solutions to the context of graphs and GNNs, since they cannot address the non-Euclidean and non-i.i.d. nature of graph learning. In this paper, we mainly focus on the graph unlearning problem, and provide innovative designs to overcome the limitations in existing solutions, as discussed in previous Section~\ref{sec:relatedwork}.

\subsubsection{\textbf{Graph Partitioning.}} Graph partitioning, the process of splitting a graph into a fixed number of shards, is a prevalent technique in graph data management. 
Generally, previous partitioning techniques strive to balance storage load and inter-server communication overhead, thereby facilitating the efficient processing of large-scale graphs stored in distributed systems~\cite{abbas2018streaming}.
These methods partition the graph into subgraphs with balanced numbers of nodes and edges to achieve load-balancing, and minimize the edge cut between subgraphs to lower the communication cost between different server machines~\cite{karypis1997metis, stanton2012streaming, zhang2017graph, xie2014distributed, lin2021large}. 
However, in line with our discussions in Section~\ref{sec:relatedwork}, such methods may lead to imbalanced sub-datasets and suboptimal sub-models due to their tendency to cluster interconnected nodes into the same subgraph. 
Furthermore, the conventional metrics of balance in graph partitioning, which focus solely on nodes or edges, do not account for the multifaceted nature of balancing the unlearning time of each subgraph GNN model. 
As discussed in Section~\ref{sec:method_part}, the unlearning time is simultaneously related to the number of both elements (i.e., nodes and edges) in each subgraph, highlighting the inadequacy of existing node-only or edge-only balance metrics. 
These limitations strongly underscore the need for a novel and systematic graph partitioning method, specifically designed for the unique demands of retraining-dependent graph unlearning.

\subsection{Limitations}
While this paper represents a significant leap forward in enhancing model utility within retraining-centric graph unlearning methods, it is still not without its limitations. Notably, there remains a substantial utility loss compared with unlearning by retraining from scratch (as shown in Table.~\ref{tab:acc}), which may impede practical applications of the proposed GraphRevoker framework. Furthermore, a more comprehensive evaluation is needed, encompassing both retraining-based and approximate graph unlearning techniques, to fully show the effectiveness of the proposed method. 

Besides, another concern arises from the potential for information leakage during the graph partitioning phase of retraining-based unlearning methods, such as GraphEraser and our own method. Since the graph partition results remain unchanged during sub-model retraining, some residual effects of the undesirable data could persist in the partitions even after unlearning, potentially leading to unforeseen consequences. Therefore, a more thorough theoretical analysis is necessary to ensure the robustness of data impact removal provided by these methods.

\end{document}